\documentclass{bmvc2k}

\title{Cascaded Boundary Regression for Temporal Action Detection}

\addauthor{~~~Jiyang Gao}{jiyangga@usc.edu}{1}
\addauthor{~~~Zhenheng Yang}{zhenheny@usc.edu}{1}
\addauthor{~~~Ram Nevatia}{nevatia@usc.edu}{1}

\addinstitution{
 Institute for Robotics and Intelligent Systems\\
 University of Southern California\\
 Los Angeles, CA, USA
}

\runninghead{}{Cascaded Boundary Regression for Temporal Action Detection}

\begin{document}

\maketitle

\begin{abstract}
Temporal action detection in long videos is an important problem. State-of-the-art methods address this problem by applying action classifiers on sliding windows. Although sliding windows may contain an identifiable portion of the actions, they may not necessarily cover the entire action instance, which would lead to inferior performance. We adapt a two-stage temporal action detection pipeline with Cascaded Boundary Regression (CBR) model. Class-agnostic proposals and specific actions are detected respectively in the first and the second stage. CBR uses temporal coordinate regression to refine the temporal boundaries of the sliding windows. The salient aspect of the refinement process is that, inside each stage, the temporal boundaries are adjusted in a cascaded way by feeding the refined windows back to the system for further boundary refinement. We test CBR on THUMOS-14 and TVSeries, and achieve state-of-the-art performance on both datasets. The performance gain is especially remarkable under high IoU thresholds, \emph{e.g.} map@tIoU=0.5 on THUMOS-14 is improved from 19.0\% to 31.0\%.

\end{abstract}

\section{Introduction}
\label{sec:intro}
Temporal action detection in long videos is an important and challenging problem, which has been receiving increasing attention recently. Given a long video, the task of action detection is to localize intervals where actions of interest take place and also predict the action categories.

Good progress has been achieved in action classification \cite{simonyan2014two, tran2015learning}, where the task is to predict action classes in "trimmed" videos. Current state-of-the-art methods \cite{oneata2014lear, Yuan_2016_CVPR, Shou_2016_CVPR} on action detection extend classification methods to detection by applying action classifiers on dense sliding windows. However, while sliding windows may contain an identifiable portion of the action, they do not necessarily cover the entire action instance or they could contain extraneous background frames, which may lead to inferior performance. Similar observations have also been made for use of sliding windows in object detection \cite{sermanet2013overfeat}. Inspired by object detection, Shou \emph{et al.} \cite{Shou_2016_CVPR} proposed a two-stage pipeline for action detection, called SCNN. In the first stage, it produces actionness scores for multi-scale sliding windows and outputs the windows with high scores as class-agnostic temporal proposals; in the second stage, SCNN categorizes the proposals to specific actions. However, SCNN still suffers from the imprecision of sliding window intervals.   

To improve temporal localization accuracy, recently a method called TURN \cite{gao2017turn} proposes to use temporal boundary regression. TURN takes sliding windows and their surrounding context as input and refine their temporal boundaries by learning a boundary regressor. We propose that the process of boundary estimation can be improved by a cascade, where a regressed clip is fed back to the system for further refinement, as the system could observe different content in each round of refinement, and refine the boundary gradually.

We adapt a two-stage action detection model with temporal coordinate regression. In the first stage, our model takes sliding windows as input, and outputs class-agnostic temporal proposals. In the second stage, our model detects actions based on the proposals. The salient aspect in our model is that, inside each stage, we propose to use Cascaded Boundary Regression (CBR) to adjust temporal boundaries in a regression cascade, where regressed clips are fed back to the system for further boundary refinement. We evaluate CBR on two challenging datasets: THUMOS-14 and TVSeries \cite{de2016online}. CBR outperforms state-of-the-art methods on both temporal action proposal generation and action detection tasks by a large margin. The performance gain is especially remarkable under high IoU thresholds, e.g map@tIoU=0.5 on THUMOS-14 is improved from 19.0\% to 31.0\%.

Our contributions are two-fold:

(1) We propose a Cascaded Boundary Regression method for temporal boundary estimation, which is shown to be effective on both proposal generation and action detection.

(2) We evaluate CBR on both proposal generation and action detection, and achieve state-of-the-art performance on both THUMOS-14 and TVSeries \cite{de2016online}.

\section{Related Work}
Temporal action detection, temporal proposal generation and object detection are related to our work, we will introduce these three topics in this section.

\textbf{Temporal Action Detection}
Temporal action localization has been received much attentions recently.  S-CNN \cite{Shou_2016_CVPR} presented action detection framework which involves two stages: the first stage uses proposal network to generate temporal action proposals; the second stage  classifies the proposals with localization network, which is trained using classification and localization loss. Singh \emph{et al.} \cite{Singh_2016_CVPR} extended two-stream \cite{simonyan2014two} framework to multi-stream network and use bi-directional LSTM networks to encode temporal information, they achieved state-of-the-art performance on MPII-Cooking dataset \cite{rohrbach2012database}. Ma \emph{et al.} \cite{Ma_2016_CVPR}  addressed the problem of early action detection. They proposed to train a LSTM network with ranking loss and merge the detection spans based on the frame-wise prediction scores generated by the LSTM. Sun \emph{et al.} \cite{sun2015temporal} proposed to transfer knowledge from web images to address temporal detection in untrimmed web videos. 

\textbf{Temporal Action Proposal.} Similar to object proposal generation, temporal proposal generation aims to produce class-agnostic proposals efficiently and accurately. Sparse-prop \cite{Heilbron_2016_CVPR} presented a method that use STIPs \cite{laptev2005space} and dictionary learning for class-independent proposal generation. SCNN-prop \cite{Shou_2016_CVPR} presented a method that fine-tunes 3D convolutional network \cite{tran2015learning} for binary proposal classification. DAPs \cite{escorcia2016daps} used LSTM networks to encode a video stream and produce proposals inside the video stream. Gao \emph{et al.} \cite{gao2017turn} proposed a method, called TURN,  to use unit-level temporal coordinate regression to refine the temporal action boundary.

\textbf{Object Proposals and Object Detection.} Recent successful object detection frameworks \cite{Girshick_2014_CVPR, girshick2015fast, ren2015faster} are built on high quality object proposals. SelectiveSearch \cite{uijlings2013selective} and Edgebox \cite{zitnick2014edge} rely on hand-crafted low-level visual features. R-CNN \cite{Girshick_2014_CVPR} and Fast R-CNN \cite{girshick2015fast} use this type of object proposals as input. RPNs \cite{ren2015faster} proposed to use anchor boxes and spatial regression for object proposal generation, which is based on ConvNet's conv-5 featmap. YOLO \cite{Redmon_2016_CVPR} proposed to divide the input image into grid cells and estimate the object bounding box by coordinate regression. SSD \cite{liu2015ssd} further adopted multi-scale grid cells to predict bounding boxes. 

\section{Methods}
\begin{figure*}[]
\centering
\includegraphics[scale=0.45]{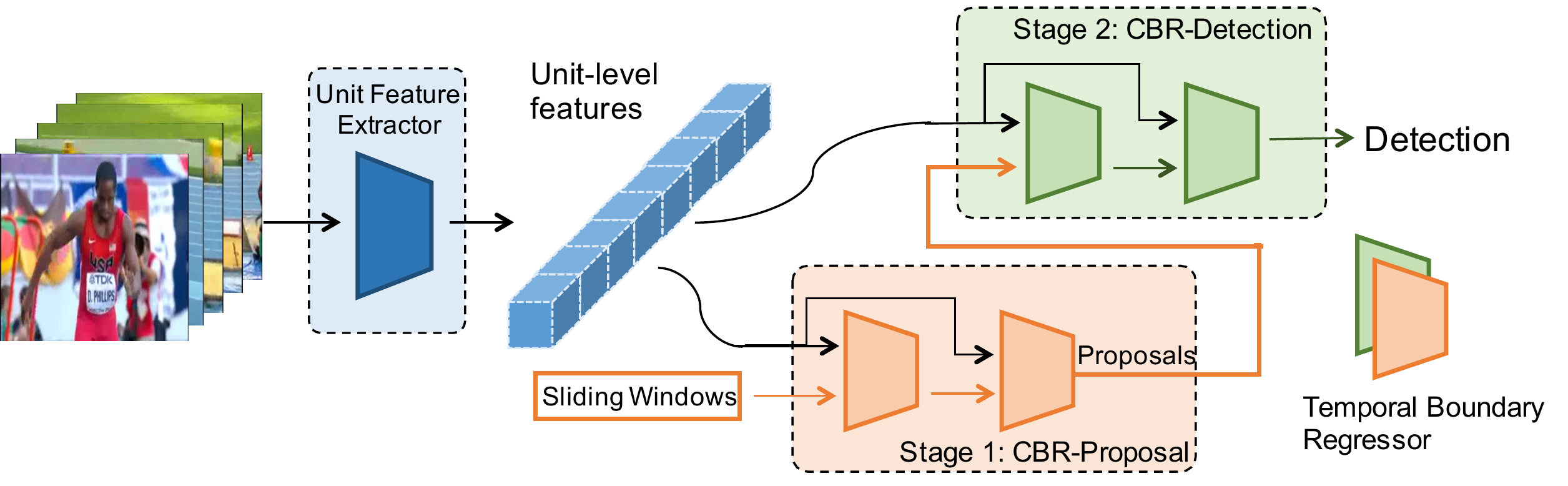}
\caption{ Architecture of two-stage action detection pipeline with Cascaded Boundary Regression (CBR)}
\label{pipeline}
\end{figure*}

In this section, we describe the two-stage Cascaded Boundary Regression (CBR) network and the training procedure, its architecture is shown in Figure \ref{pipeline}.

\subsection{Video Unit Feature Extraction} \label{sec:unit processing}
A video $V$ containing $T$ frames, $V=\{f_i\}_1^T$, is divided into $T/u_f$ consecutive video units , where $u_f$ is the number of frames in a unit. A video unit can be represented as $u=\{t_i\}_{f_s}^{f_s+u_f-1}$, where $f_s$ is the starting frame, $f_s+u_f-1$ is the ending frame $f_e$. Units do not overlap with each other. Each unit is processed by a visual encoder $E_v$ to get a unit-level representation $f_u=E_v(u)$. In our experiments, C3D \cite{tran2015learning} and two-stream CNN models \cite{simonyan2014two} are investigated. Details are given in Section 4. 

\subsection{Video Clip Modeling}
A \emph{clip} $c$ is composed of units, $c=\{u_j\}_{u_s}^{u_s+c_u-1}$, where $u_s$ is the index of starting unit and $c_u$ is the number of units inside $c$. $u_e=u_s+c_u-1$ is the index of ending unit $u_e$, and $\{u_j\}_{u_s}^{u_e}$ are called \emph{internal units} of $c$. Besides the internal units, \emph{surrounding units} for $c$ are also modeled. $\{u_j\}_{u_s-n_{ctx}}^{u_s-1}$ and  $\{u_j\}_{u_e+1}^{u_e+n_{ctx}}$ are the surrounding units before and after $c$ respectively, $n_{ctx}$ is the number of units we consider. The surrounding units provide temporal \emph{context} for clips, which are important for temporal boundary inferring. Internal feature and context features are pooled from unit-level features separately by mean pooling operation $P$. The final feature $f_c$ for a clip is the concatenation of context features and the internal feature.

\begin{equation}
f_c=P(\{u_j\}_{s_u-n_{ctx}}^{s_u}) \parallel ~P(\{u_j\}_{s_u}^{e_u})\parallel ~P(\{u_j\}_{e_u}^{e_u+n_{ctx}})
\end{equation}
where $\parallel$ represents vector concatenation. We scan a video by multi-scale temporal sliding windows. The temporal sliding windows are modeled by two parameters: window length $l_i$ and window overlap $o_i$. Note that, although multi-scale clips would have temporal overlaps, the clip-level features are computed from unit-level features, which are only calculated once.

\subsection{Temporal Coordinate Regression}
We first introduce temporal coordinate regression and then introduce the two-stage proposal and detection pipeline. Our goal is to design a method which is able to estimate the temporal boundaries of actions. For spatial boundary regression, previous work \cite{ren2015faster, girshick2015fast} uses parameterized coordinate offsets, that is, the boundary coordinates are first parameterized by the central coordinates and the size (\emph{i.e.} length and width) of the bounding box, the offsets are calculated based on these parameterized coordinates. In the temporal coordinate settings, the parameterization offsets could be represented as, 

\begin{equation}
o_x=(x^{gt}-x^{clip})/l_{clip},  \ o_l=log(l^{gt}/l^{clip})
\end{equation}
where $x$ and $l$ denote the clip's center coordinate and clip length respectively. Variables $x^{gt}, x^{clip}$ are for ground truth clip and test clip (like wise for $l$). 

Instead of using parameterization, non-parameterized offset is to use the start end end coordinates directly. Specifically, there are two levels of coordinates: frame-level and unit-level. The frame-level coordinate is the index of the frame $f_i$; the unit-level coordinate is the index of the unit $u_j$. For an action instance, the ground truth start and end coordinates $t^{gt}_s$ and $t^{gt}_e$ are usually annotated in seconds, which could be always transferred to frame-level (multiplied by FPS) $f^{gt}_s$ and $f^{gt}_e$, the unit-level ground truth coordinates are calculated by rounding:
\begin{equation}
u^{gt}_s=<f^{gt}_s/u_f>, ~~u^{gt}_e=<f^{gt}_e/u_f>
\end{equation}
where $<\cdot>$ represents rounding, $u_f$ is the frame number in a unit. The non-parameterized regression offsets are
\begin{equation}
o_s=s^{clip}-s^{gt}, ~~ o_e=e^{clip}-e^{gt}
\end{equation}
where $s^{clip}$, $e^{clip}$ are the start and end coordinates of the input clip, which could be at frame-level or unit-level. $s^{gt}$, $e^{gt}$ are the coordinates for the matched ground truth action instance. The intuition behind unit-level coordinate regression is that, as the basic unit-level features are extracted to encode $n_u$ frames, the feature may not be discriminative enough to regress the coordinates at frame-level. Comparing with frame-level regression, unit-level coordinate regression is easier to learn, though with coarser boundaries.

\subsection{Two-Stage Proposal and Detection Pipeline}

Inspired by the proposal and detection pipeline in object detection, we design a two-stage pipeline for temporal action detection, in which class-agnostic proposals  and class-specific detections are generated respectively, shown in Figure \ref{pipeline}. In both stages, temporal coordinate regression is used to infer temporal action boundaries. Sepcifically, given a clip $c=<s,e>$, it is first processed by the proposal network, which outputs two boundary regression offsets $<o_{s},o_{e}>$ and an actionness score $p_{c}$ indicating whether $c$ is an action instance. If the output $p_{c}$ is higher than a threshold $\theta$, the detection network takes $c'=<s',e'>$ (new temporal boundaries) as input and generates $n+1$ softmax scores  $p^z,x\in[1,n+1]$ and $n$ pairs of boundary offsets $<o^z_{s},o^z_{e}>, x\in [1,n]$, where $n$ is the number of action categories. 

\subsection{Cascaded Boundary Regression}
\begin{figure*}[h]
\centering
\includegraphics[scale=0.45]{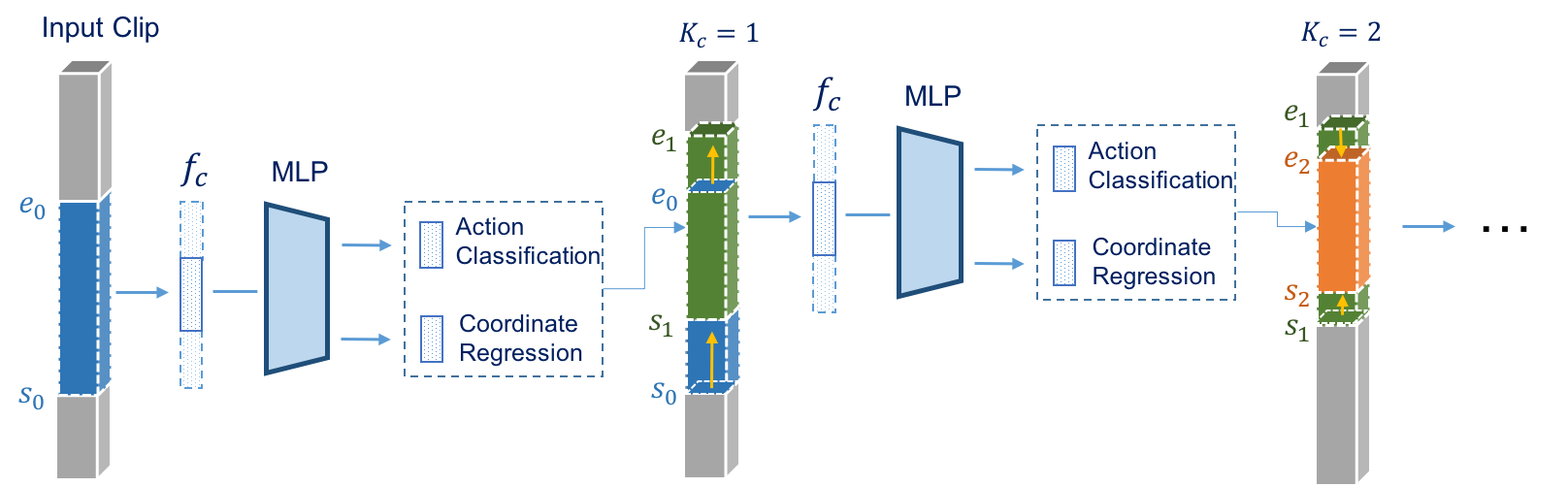}
\caption{ Unrolled model of Cascaded Boundary Regression (CBR), the parameters of the MLPs are shared.}
\label{cbr-figure}
\end{figure*}

In each stage (\emph{i.e.} the proposal and detection stages), boundary regression is applied in a cascaded manner--the output boundaries are fedback as input to the network for further refinement, as shown in Figure \ref{cbr-figure}. For proposal network, given a input clip $c=<s,e>$, the output clip $c^1=<s^1,e^1>$ is fedback as input to do a second round of refinement, and the second output is $c^2=<s^2,e^2>$. The iteration process takes $K_{c}^p$ steps, the final boundaries and the actionness score for $c$ are 
\begin{equation}
c_{K_{c}^p}=<s_{K_{c}^p},e_{K_{c}^p}>, ~~~ p=\prod_{i=1}^{K_{c}^p} p_i
\end{equation}

The cascade process of detection network is similar to that of proposal network. For detection network, it outputs $n$ pairs of temporal boundary offsets and $n+1$ category scores. Among the $n$ non-background categories, we take the category with the highest score as the prediction $x$, and pick the corresponding boundary offsets $<o_{s,x},o_{e,x}>$. The refined clip $c^1=<s^1_x,e^1_x>$ is fed back into the network. After $K_c^d$ steps, the final boundaries and score for the predicted category $x$ are
\begin{equation}
c_{K_{c}^d}=<s_{K_{c}^d}^z,e_{K_{c}^d}^z>, ~~~ p=\prod_{i=1}^{K_{c}^d} p_i^z
\end{equation}
The proposal network and the detection network are trained separately, details could be found in the next section. In either stage (proposal or detection), each cascade step could be trained separately, but here we have chosen to use the same network parameters in each step for simplicity.

\subsection{Loss Function}
To train CBR, we collect training samples from sliding windows, whose parameters (overlap and window length) will be introduced in Section 4. A class label is assigned to a sliding window if: (1) the window clip with the highest temporal Intersection over Union (tIoU) overlaps with a ground truth clip; or (2) the window clip has tIoU larger than 0.5 with any of the ground truth clips. Note that, a single ground truth clip may assign its label to multiple window clips. Negative labels are assigned to non-positive clips whose tIoU is equal to 0.0 (\emph{i.e.} no overlap) for all ground truth clips. We design a multi-task loss $L$ to jointly train classification and coordinate regression.
\begin{equation}
L=L_{cls}+\lambda L_{reg}
\end{equation}
where $L_{cls}$ is the loss for classification, which is a standard cross-entropy loss. For proposal network, $L_{cls}$ is a binary classification cross-entropy loss; for detection network, $L_{cls}$ is a stardard multi-class cross-entropy loss. $L_{reg}$ is for temporal coordinate regression and $\lambda$ is a hyper-parameter, which is set empirically. The regression loss is 
\begin{equation}
L_{reg}=\frac{1}{N}\sum_{i=1}^{N}\sum_{z=1}^{n} l_i^z[R(\hat{o}^z_{s,i}-o^z_{s,i}) +R(\hat{o}^z_{e,i}-o^z_{e,i})]
\end{equation}
where $R$ is $L1$ distance, $N$ is batch size and $n$ is the total number of categories, $l^z_i$ is the label, when the $i$th sample is from category $z$, $l^z_i=1$, otherwise, $l^z_i=0$. $\hat{o}$ is the regression estimate offset, and $o$ is the ground truth offset. For parameterized offsets, $o^z_{s,i}$ and $o^z_{e,i}$ are replaced by $o^z_{x,i}$ and $o^z_{l,i}$. 

The learning rate and batch size are set as 0.005 and 128 respectively. We use the Adam \cite{kingma2014adam} optimizer to train CBR. The ratio of sample numbers of background to non-background in a mini-batch is set to be 10 for training proposal network. For training detection network, the number background samples are equal to the average sample numbers of all categories. $\lambda$ is set to 2 for both proposal and detection network.

\section{Evaluation}
We evaluate the effectiveness of the proposed Cascaded Boundary Regression (CBR) on standard benchmarks THUMOS-14 and TVSeries for both temporal action proposal generation and action detection. 

\textbf{Unit-level Feature Extraction.} C3D unit-level features: The C3D model is pre-trained on Sports1M \cite{Karpathy_2014_CVPR}, we uniformly sample 16 frames in a unit and then input them into C3D; the output of $fc$6 layer is used as unit-level feature. Two-stream features: We use the two-stream model \cite{xiong2016cuhk} that is pre-trained on ActivityNet v1.3 training set. In each unit, the central frame is sampled to calculate the appearance CNN feature, which is the output of  "Flatten\_673" layer in ResNet \cite{he2016deep}. For the motion feature, we sample $6$ consecutive frames at the center of a unit and calculate optical flows \cite{farneback2003two} between them; these flows are then fed into the pretrained BN-Inception model \cite{xiong2016cuhk,ioffe2015batch} and the output of "global pool" layer is extracted. The motion features and the appearance features are concatenated into 4096-dimensional vectors, which are used as unit-level features.

\subsection{Experiments on THUMOS-14}
We first introduce the datasets and the experiment setup, then discuss the experimental results on THUMOS-14.

\textbf{Dataset.} The temporal action localization part of THUMOS-14 contains over 20 hours of videos from 20 sport classes. There are 200 untrimmed videos in validation set and 213 untrimmed  videos in test set. The training set of THUMOS-14 contains only trimmed videos. We train our model on the validation set and test it on the test set.

\textbf{Experimental setup.} We perform the following experiments on THUMOS-14: (1) explore components in the proposed framework: (a) parametrized offsets vs non-parameterized unit-level regression vs non-parameterized frame-level regression, (b) cascaded steps for boundary regression; (2) comparison with state-of-the-art approaches. The unit size $u_f$ is 16, the surrounding unit number $n_{ctx}$ is set to 4. The sliding window lengths and overlaps are $\{16(16),32(16),64(16),128(32),256(64),512(128)\}$, where the numbers out of brackets are lengths of sliding windows, and the numbers in brackets are the corresponding overlaps of the sliding windows.

\textbf{Temporal coordinate regression.}
To explore which type of coordinate offsets is most effective for boundary regression in temporal action detection, we test three types: (a) parameterized coordinate offsets, which are similar to the ones in object detection \cite{ren2015faster}, (b) non-parameterized frame-level coordinate offsets and (c) non-parameterized unit-level coordinate offsets. The results of temporal action detection are listed in Table \ref{offset}. The cascade step $K_c$ is set to be $1$ for both proposal stage and detection stage. Both C3D feature and Two-stream CNN feature are tested.

\begin{table}[h]\small
\centering
\caption{Comparison of different coordinate offsets on action localization (\% mAP@tIoU=0.5): parameterized, non-parameterized frame-level, non-parameterized unit-level. The performance with no boundary regression is also listed.}
\label{offset}
\begin{tabular}{l|c|c|c|c}
\hline
          & no regression & \multicolumn{1}{c|}{parameterized} & \multicolumn{1}{c|}{non-para, frame-level} & \multicolumn{1}{c}{non-para, unit-level} \\ \hline
CBR-C3D  &      16.7         &        19.4           &          18.8         &      \textbf{20.5}            \\ \hline
CBR-TS &      22.3         &      26.1        &          25.3           &   \textbf{27.7}         \\ \hline
\end{tabular}
\end{table}

We can see that all three regression offsets provide improvement over "no regression". Unit-level offsets are more effective than frame-level offsets; we think the reason is that, the features are extracted at unit level, frame-level coordinates contain redundant information, which may make the regression task more difficult. The performance of parameterized coordinate offsets is lower than that of non-parameterized unit-level offsets. We think that the reason is that unlike objects which can be re-scaled in images with camera projection, actions' time spans can not be easily re-scaled in videos, although the time spans of the same action can be varied in different videos. Therefore, "time" itself work as a standard scale for action instances.

\textbf{Cascaded boundary regression.} We explore the effects of boundary regression cascade. Cascade step $K_c^p$ and $K_c^d$ are the number of boundary regression conducted in proposal stage and detection stage respectively. The results are shown in Table \ref{cas-prop} and Table \ref{cas-det}. We investigate the cascade step with C3D feature and two-stream CNN feature. Non-parameterized unit-level coordinate offset is adopted.

\begin{table}[h]\small
\centering
\caption{Comparison of cascaded step $K_c^p = 1,2,3,4 $ for temporal proposal generation (\% AR@F=1.0) on THUMOS-14.}
\label{cas-prop}
\begin{tabular}{l|c|c|c|c}
\hline
           & $K_c^p=1$ & \multicolumn{1}{c|}{$K_c^p=2$} & \multicolumn{1}{c|}{$K_c^p=3$} & \multicolumn{1}{c}{$K_c^p=4$} \\ \hline
CBR-C3D        &   38.6  &      \textbf{39.6}      &       39.4       &     37.8      \\ \hline
CBR-TS &   42.7  &      44.5   &        \textbf{45.2}      &        44.8       \\ \hline
\end{tabular}
\end{table}

For the proposal network (shown in Table \ref{cas-prop}), we can see that cascaded boundary regression increase the performance from $42.7$ to $45.0$ for two-stream features, and from $38.6$ to $39.6$ for C3D features. When $K_c^p=3$, two-stream CBR achieves the best performance, and after the performance peak, the performance drops slightly. 

\begin{table}[h]\small
\centering
\caption{Comparison of cascaded step $K_c^d = 1,2,3,4 $ for temporal action detection (\% mAP@tIoU=0.5) on THUMOS-14.}
\label{cas-det}
\begin{tabular}{l|c|c|c|c}
\hline
           & $K_c^d=1$ & \multicolumn{1}{c|}{$K_c^d=2$} & \multicolumn{1}{c|}{$K_c^d=3$} & \multicolumn{1}{c}{$K_c^d=4$} \\ \hline
CBR-C3D        &   21.5  &      \textbf{22.7}      &       22.4       &     22.2      \\ \hline
CBR-TS &   28.4  &      \textbf{31.0}   &        30.5      &        30.2       \\ \hline
\end{tabular}
\end{table}

To test the effects of cascaded boundary regression for action detection, we fix $K_c^p=3$.  As shown in Table \ref{cas-det}, we observe a similar trend as in proposal generation: when $K_c^d=2$, CBR increases the performance from $28.4$ to $31.0$ for two-stream features, and from $21.5$ to $22.7$ for C3D features. After $K_c^d=2$, the performance becomes saturated. 

\begin{table}[h]\footnotesize
\centering
\caption{Comparison with state-of-the-art on temporal action proposal generation. Average Recall at Proposal Frequency (AR@F=1.0) performance are reported.}
\label{proposal-st}
\begin{tabular}{l|c|c|c|c||c|c}
\hline
Method & Sparse-prop\cite{Heilbron_2016_CVPR} & DAPs\cite{escorcia2016daps} & SCNN-prop\cite{Shou_2016_CVPR}  & TURN-FL \cite{gao2017turn}  & CBR-FL & CBR-TS \\ \hline
AR@AN=200 &  32.3  &  34.1    &    37.2 &  42.8   &   43.5   & \textbf{44.2}    \\   \hline 
AR@F=1.0 &  33.3  &  35.7    &    38.3 &  43.5   &   44.4   &  \textbf{45.2}  \\   \hline      
\end{tabular}
\end{table}

\begin{table}[]\small
\centering
\caption{Temporal action detection performance (mAP \%) comparison at different tIoU thresholds on THUMOS-14.}
\label{thumos-det}
\begin{tabular}{l|ccccccc}
\hline
tIoU     & 0.1           & 0.2           & 0.3           & 0.4           & 0.5   &0.6 &0.7        \\ \hline
Oneata \emph{et al.}\cite{oneata2014lear} & 36.6          & 33.6          & 27.0          & 20.8          & 14.4    &   8.5   & 3.2      \\ 
Yeung \emph{et al.}\cite{Yeung_2016_CVPR}  & 48.9          & 44.0          & 36.0          & 26.4          & 17.1    &-&-      \\ 
Yuan \emph{et al.} \cite{Yuan_2016_CVPR}  & 51.4          & 42.6          & 33.6          & 26.1          & 18.8 &-&-         \\ 
S-CNN \cite{Shou_2016_CVPR}         & 47.7          & 43.5          & 36.3          & 28.7          & 19.0    &10.3  & 5.3     \\ \hline
CBR-C3D          & 48.2      & 44.3          & 37.7          & 30.1          & 22.7  &13.8 &7.9\\ 

CBR-TS              & \textbf{60.1} & \textbf{56.7} & \textbf{50.1} & \textbf{41.3} & \textbf{31.0}  &\textbf{19.1}  &\textbf{9.9}\\ \hline
\end{tabular}
\end{table}

\textbf{Comparison with state-of-the-art on temporal proposal generation.}
We compare CBR-P with state-of-the-art methods on temporal action proposal generation, including SCNN-prop \cite{Shou_2016_CVPR}, DAPs\cite{escorcia2016daps} and Sparse-prop \cite{Heilbron_2016_CVPR} and TURN \cite{gao2017turn}. The results are shown in Table \ref{proposal-st}. To fairly compare with TURN, we also provide the performance using only optical flow CNN features, which is the same for TURN-FL. We can see that CBR-FL outperforms state-of-the-art (TURN-FL) and CBR-TS provides further improvement over CBF-FL.

\textbf{Comparison with state-of-the-art on temporal action detection.}
We compare our method with other state-of-the-art temporal action localization methods on THUMOS-14, the results are shown in Table \ref{thumos-det}. We compare with the challenge results \cite{oneata2014lear}, and recent methods including based on segment window C3D \cite{Shou_2016_CVPR}, score pyramids \cite{Yuan_2016_CVPR} and deep recurrent reinforcement learning \cite{Yeung_2016_CVPR}. Both SCNN and CBR-C3D are based on C3D features, we can see that CBR-C3D outperforms SCNN at all tIoU thresholds, especially at high tIoU, which shows the effectiveness of CBR. If two-stream features are adopted, CBR outperforms state-of-the-art methods by 12\% at tIoU=0.5.

\subsection{Experiments on TVSeries}
We first introduce the datasets and the experiment setup, then discuss the experimental results on TVSeries \cite{de2016online}.

\textbf{Dataset.} The TVSeries Dataset \cite{de2016online} is a realistic, large-scale dataset for temporal action detection, which contains 16 hours of videos (27 episodes) from six recent popular TV series. 30 daily life action categories are defined in TVSeries, such as "close door", "drive car", "wave". There are totally 6231 action instances annotated with start and end times and action categories in the dataset. The train/validation/test sets contain 13/7/7 episodes respectively.

\textbf{Experimental setup.} We test the cascaded steps of boundary regression for both proposal generation and action detection, and then compare with state-of-the-art performance on action detection. The unit size $u_f$ is 6, the surrounding unit number $n_{ctx}$ is set to 4. The sliding window lengths and overlaps are $\{12(6),24(6),48(12),72(18),96(24),192(48),384(96)\}$, where the numbers out of brackets are lengths of sliding window, and the numbers in brackets are the corresponding overlaps of the sliding window.

\textbf{Cascaded boundary regression.}
We explore the effects of cascaded boundary regression on TVSeries. Note that $K_c$ is the cascaded step. We investigate the cascaded step with two-stream CNN features.

\begin{table}[h]\small
\centering
\caption{Comparison of cascaded step $K_c = 0,1,2,3,4 $ for temporal action detection (\% mAP@tIoU=0.5) on TVSeries. $K_c = 0$ means that the system only do classification, no boundary regression.}
\label{cas-tv}
\begin{tabular}{l|c|c|c|c|c}
\hline
           & \multicolumn{1}{c|}{$K_c=0$} & $K_c=1$ & \multicolumn{1}{c|}{$K_c=2$} & \multicolumn{1}{c|}{$K_c=3$} & \multicolumn{1}{c}{$K_c=4$} \\ \hline
Proposal (AR@F=1.0)        & 20.4 &   24.3  &      25.6      &       \textbf{26.1}       &     25.9      \\ \hline
Detection (mAP@IoU=0.2) & 6.2 &   8.8  &      \textbf{9.5}   &        9.2      &        9.0       \\ \hline
\end{tabular}
\end{table}

As shown in Table \ref{cas-tv}, comparing with $K_c=0$ and $K_c=1$, we can see that temporal coordinate regression brings a big improvement, which shows its effectiveness. We can also see that when $K_c=3$, CBR achieves the best performance for proposal network. To test detection network, we fix $K_c^p=3$, the results show that when  $K_c^d=2$ CBR achieves the best performance for action detection. After the peak, the performance starts to decrease. The performance distribution of cascaded step is consistent with THUMOS-14.

\textbf{Comparison with state-of-the-art on action detection.} We compare CBR with state-of-the-art performance on TVSeires in Table \ref{tv-det}. Overall, we can see that TVSeries is a more challenging dataset than THUMOS-14. To provide another comparison, we train SVM classifiers based on two-stream features, which is shown in Table \ref{tv-det} as \textbf{SVM-TS}. The SVM classifiers are trained and tested using the same samples as CBR, which are described in Section 3.6; clip-level features are mean-pooled from unit-level features. We can see that with the same features, CBR outperforms the SVM-based classifiers by 3.5\% at tIoU=0.2.  At $tIoU=0.2$, CBR achieves 9.5, while the state-of-the-art method FV \cite{de2016online} only achieves $4.9$. We also report mAP performance at $tIoU=0.1$ and $tIoU=0.3$, which are 11.0 and 7.9 respectively.

\begin{table}[h]\small
\centering
\caption{Temporal action detection performance (mAP \%) comparison at different tIoU thresholds on TVSeries.}
\label{tv-det}
\begin{tabular}{l|ccccc}
\hline
tIoU     & CNN \cite{de2016online}           & LSTM \cite{de2016online}           & FV  \cite{de2016online}     & SVM-TS & CBR-TS         \\ \hline
 0.1 & -          & -          & -     & 7.3&       \textbf{11.0}     \\ 
 0.2 &1.1          & 2.7          & 4.9   & 6.0 &      \textbf{9.5}     \\ 
  0.3 & -          & -          & -      &4.6 &      \textbf{7.9}    \\ \hline
\end{tabular}
\end{table}

\section{Conclusion}
We present a novel two-stage action detection pipeline with Cascaded Boundary Regression (CBR), which achieves state-of-the-art performance on standard benchmarks. In the first stage, temporal proposals are generated; based on the proposals, actions are detected in the second stage. Cascaded boundary regression are conducted in both stages. Detailed experiments and analysis on cascaded steps are conducted, which show the effectiveness of CBR for both temporal proposal generation and action detection. Different temporal regression offset settings are also investigated and discussed. State-of-the-art performance has been achieved on both THUMOS-14 and TVSeires dataset.

\bibliography{bmvc_review}
\end{document}